\documentclass[10pt,twocolumn,twoside]{IEEEtran}

\usepackage{tcolorbox}
\usepackage{graphicx}
\usepackage{xcolor}
\usepackage{theorem}
\usepackage{times,amsmath,epsfig}
\usepackage{amssymb}
\usepackage{subfigure}
\usepackage{cite}
\usepackage{url}
\usepackage{multicol}
\usepackage{bm}

\usepackage[ruled,vlined]{algorithm2e}
\usepackage{hyphenat}
\input{mysymbol.sty}

\newtheorem{mytheorem}{\bf Theorem}

\newtheorem{mylemma}{\bf Lemma}
\newtheorem{myproposition}{\bf Proposition}

\newtheorem{problem}{\bf Problem}

\title{Accelerated Graph Learning from Smooth Signals}

\author{\IEEEauthorblockN{Seyed Saman Saboksayr, \IEEEmembership{Student Member, IEEE}, and Gonzalo Mateos, \IEEEmembership{Senior Member, IEEE}}
\thanks{Paper submitted on \today. Work in this paper was supported in part by the NSF awards CCF-1750428, CCF-1934962 and ECCS-1809356. The authors are with the Dept. of Electrical and Computer Eng., University of Rochester. Emails: ssaboksa@ur.rochester.edu and gmateosb@ece.rochester.edu.}} 

\begin{document}
\maketitle

\begin{abstract}
We consider network topology identification subject to a signal smoothness prior on the nodal observations. A fast dual-based proximal gradient algorithm is developed to efficiently tackle a strongly convex, smoothness-regularized network inverse problem known to yield high-quality graph solutions. Unlike existing solvers, the novel iterations come with global convergence rate guarantees and do not require additional step-size tuning. Reproducible simulated tests demonstrate the effectiveness of the proposed method in accurately recovering random and real-world graphs, markedly faster than state-of-the-art alternatives and without incurring an extra computational burden.
\end{abstract}



\section{Introduction}\label{S:Introduction}


\IEEEPARstart{N}{etwork}-aware signal and information processing is having a major impact in technology and the biobehavioral sciences; see e.g,~\cite[Ch. 1]{kolaczyk09}. In this context, graph signal processing (GSP) builds on a graph-theoretic substrate to effectively model signals with complex relational structures~\cite{ortega18,shuman13,SandryMouraSPG_TSP13}. However, the required connectivity information is oftentimes not explicitly available. This motivates the prerequisite step of using signals (e.g., brain activity traces, distributed sensor measurements) to unveil latent network structure, or,  to construct discriminative graph representations to facilitate downstream learning tasks. As graph data grow in size and complexity, there is an increasing need to develop customized, fast and computationally-efficient graph learning algorithms.

Given nodal measurements (known as graph signals in the GSP parlance), the network topology inference problem is to search for a graph within a model class that is optimal in some application-specific sense, e.g.,~\cite[Ch. 7]{kolaczyk09}. The adopted criterion is naturally tied to the signal model relating the observations to the sought network, which can include constraints motivated by physical laws, statistical priors, or, explainability goals. Workhorse probabilistic graphical models include Gaussian Markov random fields, and topology identification arises with so-termed high-dimensional graphical model selection~\cite{dempster_cov_selec,yuanlin2007,glasso2008,Lake10discoveringstructure,egilmez2017jstsp,pavez2018tsp,kumar2020jmlr}. Other recent approaches embrace a signal representation perspective to reveal parsimonious data signatures with respect to the underlying graph. These include stationarity induced via linear network diffusion~\cite{segarra2016topoidTSP16,pasdeloup2016inferenceTSIPN16,rasoul20} and smoothness (i.e., bandlimitedness)~\cite{kalofolias16,dong16,kalofolias17,kalofolias2019iclr,sundeep_icassp17,mike_icassp17,sardellitti19,berger2020graphlearning,bars19}. The interested reader is referred to~\cite{mateos19,dong2019learning,giannakis18} for comprehensive tutorial treatments of network topology inference advances.

In this short letter, we develop a fast and scalable algorithm to estimate graphs subject to a smoothness prior (Section \ref{S:Preliminaries} outlines the required background and formally states the problem). Adopting the well-appreciated graph learning framework of~\cite{kalofolias16,kalofolias2019iclr}, in Section \ref{S:Accelerated} we bring to bear the fast proximal-gradient (PG) iterations in~\cite{beck2014} to solve the resulting strongly convex, signal smoothness-regularized optimization problem in the dual domain. There are noteworthy recent scalable solvers for this problem that rely on the primal-dual (PD) method~\cite{kalofolias16}, PG~\cite{saboksayr21eusipco_ogl}, or, the linearized alternating-direction method of multipliers (ADMM)~\cite{wang2021}. Unlike these algorithms, the novel iterations come with global convergence rate guarantees and do not require additional step-size tuning. Borrowing results from~\cite{beck2014}, we show that a (possibly infeasible) primal sequence generated from the accelerated graph learning algorithm converges to a globally optimal solution at a rate of $O(1/k)$. To the best of our knowledge, this is the first work that establishes the convergence rate of topology inference algorithms subject to smoothnes priors. Computer simulations in Section \ref{S:Simulations} showcase the favorable convergence properties of the proposed approach when recovering a wide variety of graphs. In the interest of reproducible research, the code used to generate all figures in this letter is publicly available. Conclusions are in Section \ref{S:conclusions}. Due to page contraints, proofs are deferred to the accompanying Supplementary Material.


\section{Graph Learning from Smooth Signals}
\label{S:Preliminaries}


Let $\ccalG \left ( \ccalV, \ccalE, \bbW \right)$ be an undirected graph, where $\ccalV$ are the nodes (or vertices) with $|\ccalV|=N$, $\ccalE \subseteq \ccalV \times \ccalV$ are the edges, and $\bbW \in \reals^{N \times N}_{+}$ is the symmetric adjacency matrix collecting the edge weights. For $\left (i,j \right ) \notin \ccalE$ we have $W_{ij}=0$. We exclude the possibility of self-loops, so $\bbW$ is hollow meaning $W_{ii}=0$, $\forall \:i\in\ccalV$. We acquire graph signal observations $\bbx =\left[x_1,\dots,x_N\right]^{\top}\in\reals^N$, where $x_i$ denotes the signal value at vertex $i \in \ccalV$. More general graphs capturing directionality are important~\cite{marques20}, but beyond the scope of this letter.


\subsection{Graph signal smoothness}\label{Ss:smoothness}


For undirected graphs one typically adopts the Laplacian $\bbL := \diag \left( \bbd \right) - \bbW$ as descriptor of graph structure, where $\bbd=\bbW\mathbf{1}$ collects the vertex degrees. 
As the central object in spectral graph theory, $\bbL$ is instrumental in formalizing the notion of smooth (i.e., low-pass bandlimited) signals on graphs~\cite{zhou04,ortega18}. Specifically, the total variation (TV) of the graph signal $\bbx$ with respect to $\ccalG$ is given by the quadratic form
\begin{equation}\label{eq5}
	\textrm{TV}(\bbx):=\bbx^{\top}\bbL \bbx
	= \frac{1}{2}\sum_{i \neq j} W_{ij} \left( x_i - x_j \right)^2.
\end{equation}
We can interpret $\textrm{TV}(\bbx)$ as a smoothness measure for graph signals, which gauges the extent to which $\bbx$ varies across local neighborhoods in $\ccalG$. Accordingly, we say a signal is smooth if it has a small total variation. For reference, $0\leq \textrm{TV}(\bbx)\leq \lambda_{\max}$, where $\lambda_{\max}$ is the spectral radius of $\bbL$. The lower bound is attained by constant signals. The ubiquity of smooth network data has been well-documented, with examples spanning sensor measurements~\cite{chepuri17}, protein function annotations~\cite{kolaczyk09}, and product ratings~\cite{weiyu2018tsp}. These empirical findings motivate adopting smoothness as the criterion to search for graphs on which measurements exhibit desirable parsimony or regularity.\vspace{2pt}


\subsection{Problem statement}\label{Ss:problem}


We study the following graph learning problem.\vspace{-5pt}
\begin{problem}\label{P:problem_statement}
	Given a set $\ccalX:=\{\bbx_p\}_{p=1}^P$ of graph signal observations, the goal is to learn an undirected graph $\ccalG(\ccalV,\ccalE, \bbW)$ such that the observations in $\ccalX$ are smooth on $\ccalG$.
\end{problem}
We now briefly review the method proposed in~\cite{kalofolias16,kalofolias2019iclr} to tackle Problem \ref{P:problem_statement}, from which we henceforth build on to develop a fast graph learning algorithm.

Consider the matrix $\bbX=[\bbx_1,\ldots,\bbx_P]\in \reals^{N\times P}$, whose columns $\bbx_p$ are the observations in $\ccalX$. The rows,  denoted by $\bar{\bbx}_i^{\top}\in\reals^{1\times P}$, collect all $P$ measurements at vertex $i$. Define then the nodal Euclidean-distance matrix $\bbE\in\reals_{+}^{N\times N}$, where $E_{ij}:=\|\bar{\bbx}_i-\bar{\bbx}_j\|_2^2$, $i,j\in\ccalV$. Using these notions, the signal smoothness measure over $\ccalX$ can be equivalently written as
\begin{equation}\label{E:smooth_sparse}
	\sum_{p=1}^P\textrm{TV}(\bbx_p)=\textrm{trace}(\bbX^{\top}\bbL\bbX)=\frac{1}{2}\|\bbW\circ\bbE\|_1,
\end{equation}
where $\circ$ denotes element-wise product~\cite{kalofolias16}. Smoothness minimization as criterion in Problem \ref{P:problem_statement} has the following intuitive interpretation: when pairwise nodal distances in $\bbE$ are sampled from a smooth manifold, the learnt topology $\bbW$ tends to be sparse, preferentially choosing edges $(i,j)$ whose corresponding $E_{ij}$ are smaller [cf. the weighted $\ell_1$-norm in \eqref{E:smooth_sparse}]. 

Leveraging this neat link between signal smoothness and edge sparsity, a fairly general graph-learning framework was put forth in~\cite{kalofolias16}. The idea therein is to solve the following convex inverse problem
\begin{align}\label{eq:kalofolias}
	\min_{\bbW}&{}\:\left\{\|\bbW\circ\bbE\|_1-\alpha\bbone^{\top} \log \left( \bbW\bbone \right)+\frac{\beta}{2}\|\bbW\|_F^2\right\}\\
	\textrm{ s. to } &{} \quad\textrm{diag}(\bbW)=\mathbf{0},\: W_{ij}=W_{ji}\geq 0, \:i\neq j\nonumber
\end{align}
where $\alpha,\beta>0$ are tunable regularization parameters. Different from~\cite{dong16}, the logarithmic barrier on the vertex degrees $\bbd=\bbW\bbone$ excludes the possibility of having (often undesirable) isolated vertices in the estimated graph. Through $\beta$, the Frobenius-norm penalty offers a handle on the graphs' edge sparsity level. Among the parameterized familiy of solutions to \eqref{eq:kalofolias}, the sparsest graph is obtained when $\beta=0$.

Arguably, the most important upshot of identity \eqref{E:smooth_sparse} is computational. It facilitates formulating \eqref{eq:kalofolias} as a search over adjacency matrices, and the resulting constraints (null diagonal, symmetry and non-negativity) are separable across the variables $W_{ij}$. This does not hold for the Laplacian $\bbL$.
Exploting this favorable structure of \eqref{eq:kalofolias}, efficient solvers were developed based on PD iterations~\cite{kalofolias16}, the PG method~\cite{saboksayr21eusipco_ogl}, or the ADMM~\cite{wang2021}. However, none of these graph learning methods come with convergence rate guarantees because the objective function of \eqref{eq:kalofolias} lacks a Lipschitz continuous gradient. To close this gap, next we develop a markedly faster first-order algorithm using an accelerated dual-based PG method~\cite{beck2014}.


\section{Fast Dual Proximal Gradient Algorithm}\label{S:Accelerated}


Because $\bbW$ is hollow and symmetric, the optimization variables in \eqref{eq:kalofolias} are effectively the, say, upper-triangular elements $[\bbW]_{ij}$, $j>i$. Thus, it suffices to retain only those entries in the vector $\bbw:=\textrm{vec}[\textrm{triu}[\bbW]] \in \reals_{+}^{N(N-1)/2}$, were we have adopted convenient Matlab notation. To impose that edge weights are non-negative, we penalize the cost with the indicator function $\ind{\bbw\succeq\mathbf{0}}=0$ if $\bbw\succeq \mathbf{0}$, else $\ind{\bbw\succeq \mathbf{0}}=\infty$~\cite{kalofolias16}. This way, we equivalently reformulate \eqref{eq:kalofolias} as the unconstrained, non-differentiable problem
\begin{equation}\label{eq:kalofolias_vec}
	\min_{\bbw}\Big\{\underbrace{\mbI\left\{\bbw\succeq\mathbf{0}\right\} + 2\bbw^{\top}\bbe+\beta\|\bbw\|_2^2}_{:=f(\bbw)}-\underbrace{\alpha \bbone^{\top} \log \left( \bbS\bbw \right)}_{ :=-g(\bbS\bbw)}\Big\}, 
\end{equation}
where $\bbe:=\textrm{vec}[\textrm{triu}[\bbE]] $ and $\bbS\in\{0,1\}^{N\times N(N-1)/2}$ maps edge weights to nodal degrees, i.e., $\bbd=\bbS\bbw$. The non-smooth function $f(\bbw):=\mbI\left\{\bbw\succeq\mathbf{0}\right\} + 2\bbw^{\top}\bbe+\beta\|\bbw\|_2^2$ is strongly convex with strong convexity parameter $2\beta$ (details are in the Supplementary Material), while $g(\bbw):=-\alpha \bbone^{\top} \log \left(\bbw \right)$ is a (strictly) convex function for all $\bbw \succ \mathbf{0}$. Under the aforementioned properties of $f$ and $g$, the composite problem \eqref{eq:kalofolias_vec} has a unique optimal solution $\bbw^\star$; see e.g.,~\cite{beck2014} and~\cite{wang2021}. 

A fast dual-based PG algorithm was developed in~\cite{beck2014} to solve the non-smooth, strictly convex optimization problem $\min_{\bbw} \big\{f(\bbw)+g(\bbS\bbw)\big\}$ of which \eqref{eq:kalofolias_vec} is a particular instance. In the remainder of this section we will bring to bear this optimization framework to develop a novel graph learning algorithm with global rate of convergence guarantees.


\subsection{The dual problem}\label{Ss:dual_problem}


The structure of \eqref{eq:kalofolias_vec} lends itself naturally to variable-splitting via the equivalent
linearly-constrained form
\begin{equation}\label{eq:primal_split}
	\min_{\bbw,\bbd}\left\{f(\bbw)+g(\bbd)\right\},\quad  \textrm{ s. to }\bbd=\bbS\bbw.
\end{equation}
Attaching Lagrange multipliers $\bblambda\in \reals^N$ to the equality constraints and minimizing the Lagrangian function $\ccalL(\bbw,\bbd,\bblambda)=f(\bbw)+g(\bbd)-\langle\bblambda,\bbS\bbw-\bbd\rangle$ w.r.t. the primal variables $\{\bbw,\bbd\}$, one arrives at the (minimization form) dual problem~\cite{beck2014}
\begin{equation}\label{eq:dual}
	\min_{\bblambda}\left\{F(\bblambda)+G(\bblambda)\right\},
\end{equation}
where
\begin{align}
	F(\bblambda):={}&\max_{\bbw}\left\{\langle \bbS^\top \bblambda,\bbw\rangle-f(\bbw)\right\}\label{eq:conjugate_F},\\
	G(\bblambda):={}&\max_{\bbd}\left\{\langle -\bblambda,\bbd\rangle-g(\bbd)\right\}.\label{eq:conjugate_G}
\end{align}
Interestingly, the strong convexity of $f$ induces useful smoothness properties for $F$ (namely, the composition of $\bbS \bbw$ with the Fenchel conjugate of $f$), that we summarize next. The result is adapted from~\cite[Lemma 3.1]{beck2014} and the additional proof arguments can be found in the Supplementary Material.\vspace{-5pt}
\begin{mylemma}\label{lemma:Lipschitz_gradient}
Function $F(\bblambda)$ in \eqref{eq:conjugate_F} is smooth, and the gradient $\nabla F(\bblambda)$ is Lipschitz continuous with constant $L:=\frac{N-1}{\beta}$.	
\end{mylemma}
This additional structure of \eqref{eq:dual} makes it feasible to apply accelerated PG algorithms~\cite{beck18} (such as FISTA~\cite{beck2009}), to solve the dual problem.


\subsection{Accelerated dual proximal gradient algorithm}\label{Ss:accelerated}


The FISTA algorithm applied to the dual problem \eqref{eq:dual} yields the following iterations (initialized as $\bbomega_{1}=\bblambda_0\in \reals^N$ and $t_1=1$, henceforth $k=1,2,\ldots$ denotes the iteration index)
\begin{align}
\bblambda_k={}&\textbf{prox}_{L^{-1}G}\left(\bbomega_{k} - \frac{1}{L} \nabla F(\bbomega_{k}) \right)\label{eq:FISTA_prox},\\
t_{k+1}={}&\frac{1+\sqrt{1+4t_{k}^2}}{2},\label{eq:FISTA_t}\\
\bbomega_{k+1}={}& \bblambda_k+\left(\frac{t_k-1}{t_{k+1}}\right)\left[ \bblambda_k- \bblambda_{k-1}\right],	\label{eq:FISTA_extrapolation}
\end{align}
where the proximal operator of a proper, lower semi-continuous convex function $h$ is (see e.g.,~\cite{boyd14})
\begin{equation}\label{eq:prox_operator}
\textbf{prox}_{h}(\bbx)=\argmin_{\bbu}\left\{h(\bbu)+\frac{1}{2}\|\bbu-\bbx\|_2^2\right\}.
\end{equation}
An adaptation of the result in~\cite[Lemma 3.2]{beck2014} -- stated as Proposition \ref{prop:iterations} below -- yields the novel graph learning iterations tabulated under Algorithm \ref{A:accelerated}. Again, due to page constraints the proof details are deferred to the Supplementary Material.\vspace{-5pt}
\begin{myproposition}\label{prop:iterations}
The dual variable update iteration in \eqref{eq:FISTA_prox} can be equivalently rewritten as $\bblambda_k=\bbomega_k-L^{-1}(\bbS\bar{\bbw}_k-\bbu_k)$, with
\begin{align} 	
\bar{\bbw}_k={}& \max\left(\mathbf{0},\frac{\bbS^\top\bbomega_k-2\bbe}{2\beta}\right)\label{eq:barw_update},\\ 
\bbu_k={}&\frac{\bbS\bar{\bbw}_k-L\bbomega_k  + \sqrt{(\bbS\bar{\bbw}_k-L\bbomega_k)^2 + 4\alpha L\bbone}}{2},\label{eq:u_update}
\end{align}	
where $\max(\cdot,\cdot)$ in \eqref{eq:barw_update} as well as both $(\cdot)^2$ and $\sqrt{(\cdot)}$ in \eqref{eq:u_update} are element-wise operations on their vector arguments.
\end{myproposition}
The updates in Proposition \ref{prop:iterations} are fully expressible in terms of parameters from the original graph learning problem, namely $N$, $\alpha$, $\beta$, $\bbS$ and the data in $\bbe$. This is to be contrasted with \eqref{eq:FISTA_prox}, which necessitates the conjugate functions $F$ and $G$. 

Algorithm \ref{A:accelerated}'s overall computational complexity is dominated by the update \eqref{eq:barw_update}, which incurs a per iteration cost of $\ccalO(N^2)$. The remaining updates are also given in closed form, through simple operations of vectors living in the dual $N$-dimensional domain of nodal degrees [cf. the $N(N-1)/2$-dimensional primal variables $\bar{\bbw}_k$]. The overall complexity of $O(N^2)$ is in par with state-of-the-art PD and linearized ADMM algorithms~\cite{wang2021}, which have been shown to scale well to large networks with $N$ in the order of thousands. The computational cost can be further reduced by constraining a priori the space of possible edges; see~\cite{kalofolias2019iclr} for examples where this approach is warranted. For a given problem instance, there are no step-size parameters to tune here (on top of $\alpha$ and $\beta$) since we can explicitly compute the Lipschitz constant $L$ in Lemma \ref{lemma:Lipschitz_gradient}. On the other hand, the linearized ADMM algorithm in~\cite{wang2021} necessitates tuning two step-sizes and the penalty parameter defining the augmented Lagrangian. 

The distinctive feature of the proposed accelerated dual PG algorithm is that it comes with global convergence rate guarantees. These results are outlined in the ensuing section.	


\begin{algorithm}[t]\label{A:accelerated}
	\SetAlgoLined
	\textbf{Input} parameters $\alpha,\beta$, data $\bbe$, set $L=\frac{N-1}{\beta}$.\\
	\textbf{Initialize} $t_1=1$ and $\bbomega_{1}=\bblambda_0$ at random. \\
	\For{$k=1,2,\dots,$}{
		$\bar{\bbw}_k= \max\left(\mathbf{0},\frac{\bbS^\top\bbomega_k-2\bbe}{2\beta}\right)$\\
		$\bbu_k=\frac{\bbS\bar{\bbw}_k-L\bbomega_k  + \sqrt{(\bbS\bar{\bbw}_k-L\bbomega_k)^2 + 4\alpha L\bbone}}{2}$\\
		$\bblambda_k=\bbomega_k-L^{-1}(\bbS\bar{\bbw}_k-\bbu_k)$\\
		$t_{k+1}=\frac{1+\sqrt{1+4t_{k}^2}}{2}$\\
		$\bbomega_{k+1}=\bblambda_k+\left(\frac{t_k-1}{t_{k+1}}\right)\left[ \bblambda_k- \bblambda_{k-1}\right]$
	}
	\textbf{Output} graph estimate $\hbw_k=\max\left(\mathbf{0},\frac{\bbS^\top\bblambda_k-2\bbe}{2\beta}\right)$
	\caption{Topology inference via fast dual PG (FDPG)}
\end{algorithm}



\subsection{Convergence rate analysis}\label{Ss:convergence_rate}


Moving on to convergence properties, when $k\to\infty$ the iterates $\bblambda_k$ generated by Algorithm \ref{A:accelerated} provably approach a dual optimal solution $\bblambda^\star$ that minimizes $\varphi(\bblambda):=F(\bblambda)+G(\bblambda)$ in \eqref{eq:dual}; see e.g.,~\cite{beck2009}. The celebrated FISTA rate of convergence for the dual cost function is stated next. \vspace{-5pt}
\begin{mytheorem}{\normalfont\cite[Theorem 4.4]{beck2009}}\label{th:FISTA_rate_dual_cost}
For all $k\geq 1$, dual iterates $\bblambda_{k}$ stemming from Algorithm \ref{A:accelerated} are such that
\begin{equation}\label{eq.th1}
\varphi(\bblambda_{k})-\varphi(\bblambda^\star)\leq \frac{2(N-1)\|\bblambda_0-\bblambda^\star\|_2^2}{\beta k^2}.
\end{equation}
\end{mytheorem}
This well-documented $O(1/k^2)$ global convergence rate of accelerated PG algorithms implies an $\ccalO(1/\sqrt{\epsilon})$ iteration complexity to return an $\epsilon$-optimal dual solution measured in terms of $\varphi$ values.  

We now consider a primal sequence generated from the iterates of Algorithm \ref{A:accelerated}, and borrow the results from~\cite{beck2014} to show the sequence is globally convergent to $\bbw^\star$ at a rate of $O(1/k)$. To this end, suppose that for all $k\geq 1$ we are given dual updates $\bblambda_k$ generated from the accelerated dual PG algorithm. We can construct a primal sequence as $\hbw_k=\argmin_{\bbw}\ccalL(\bbw,\bbd,\bblambda_k)$, namely [cf. \eqref{eq:conjugate_F}]
\begin{align}
\hbw_k=&{}\argmax_\bbw\left\{\langle \bbS^\top \bblambda_k,\bbw\rangle-f(\bbw)\right\}\nonumber\\
=&{}\max\left(\mathbf{0},\frac{\bbS^\top\bblambda_k-2\bbe}{2\beta}\right).\label{eq:primal_seq}
\end{align}
As noted in~\cite{wang2021}, this primal sequence may be infeasible in the sense that resulting nodal degrees $\hbd_k:=\bbS\hbw_k$ are not guaranteed to lie in the domain of $g$. The promised $O(1/k)$ rate of converge result for $\hbw_k$ is stated next.\vspace{-5pt}
\begin{mytheorem}{\normalfont\cite[Theorem 4.1]{beck2014}}\label{th:FISTA_rate_primal_var}
For all $k\geq 1$, the primal sequence \eqref{eq:primal_seq} defined in terms of dual iterates $\bblambda_{k}$ generated by Algorithm \ref{A:accelerated} satisfies
\begin{equation}\label{eq.th1}
\|\hbw_k-\bbw^\star\|_2\leq \frac{\sqrt{2(N-1)}\|\bblambda_0-\bblambda^\star\|_2}{\beta k}.
\end{equation}
\end{mytheorem}
%


\section{Numerical Results}\label{S:Simulations}


\begin{figure*}[!t]
    \centering
    \begin{minipage}[c]{\textwidth}
    \centering
    \includegraphics[width=0.8\textwidth]{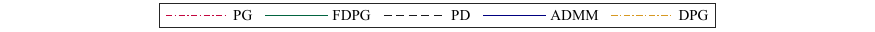}
    \end{minipage}
    \vspace{-8pt}
    \newline
    \begin{minipage}[c]{.32\textwidth}
    \begin{minipage}[c]{.49\textwidth}
    \includegraphics[width=\textwidth]{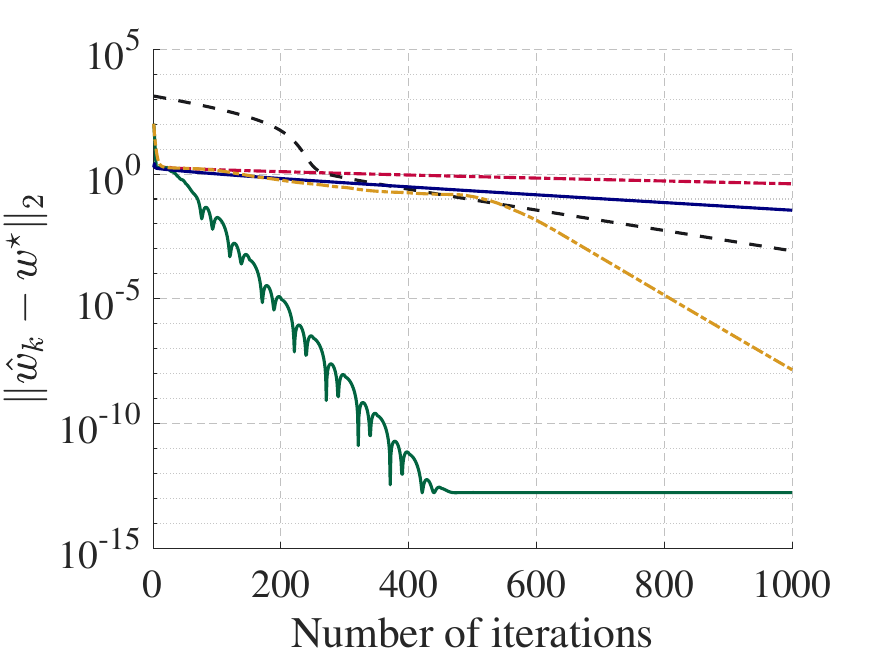}
    \end{minipage}
    \begin{minipage}[c]{.49\textwidth}
    \includegraphics[width=\textwidth]{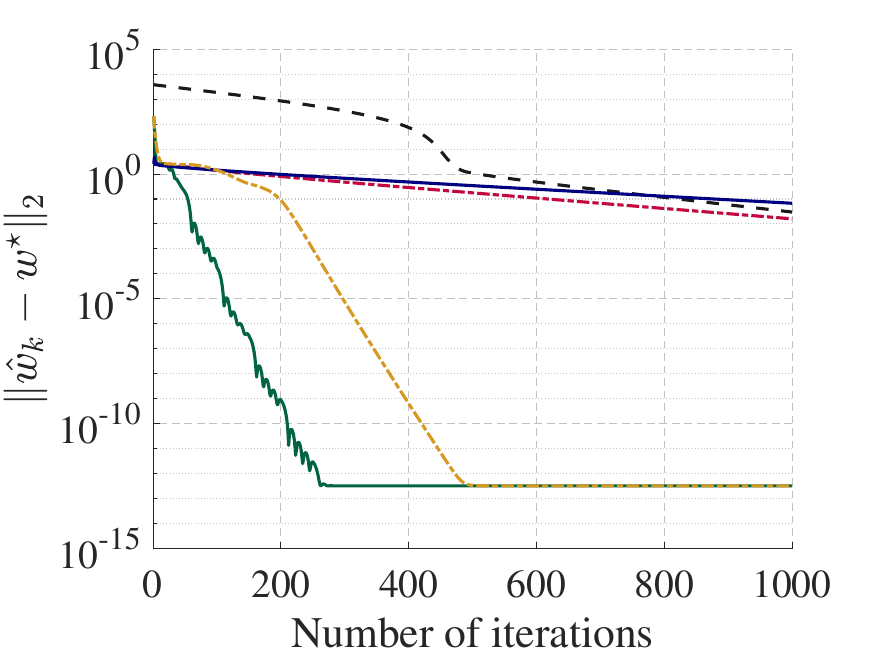}
    \end{minipage}
    \begin{minipage}[c]{.49\textwidth}
    \includegraphics[width=\textwidth]{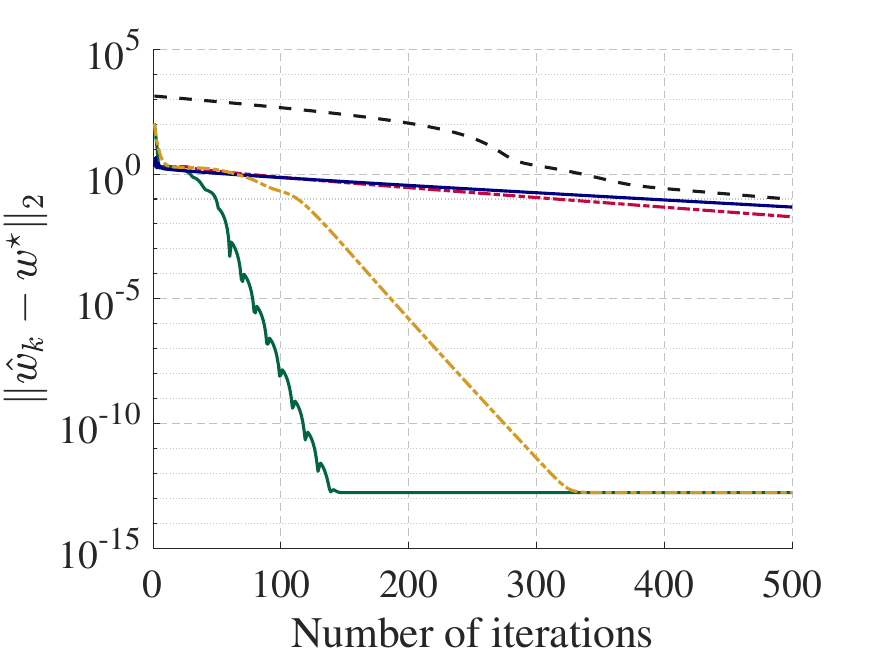}
    \end{minipage}
    \begin{minipage}[c]{.49\textwidth}
    \includegraphics[width=\textwidth]{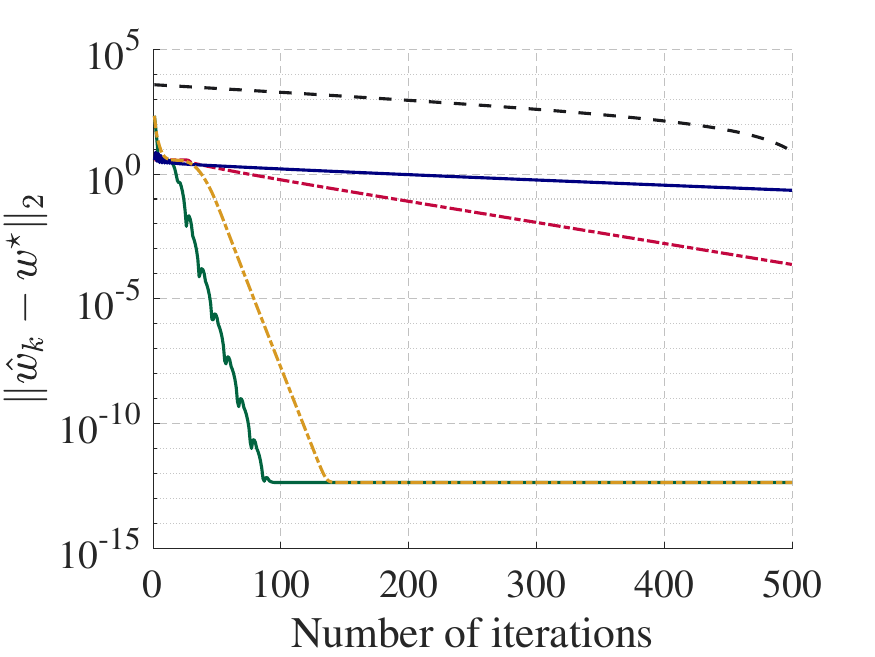}
    \end{minipage}
    \newline
    \centering{\small (a)}
    \end{minipage}
    \begin{minipage}[c]{.32\textwidth}
    \begin{minipage}[c]{.49\textwidth}
    \includegraphics[width=\textwidth]{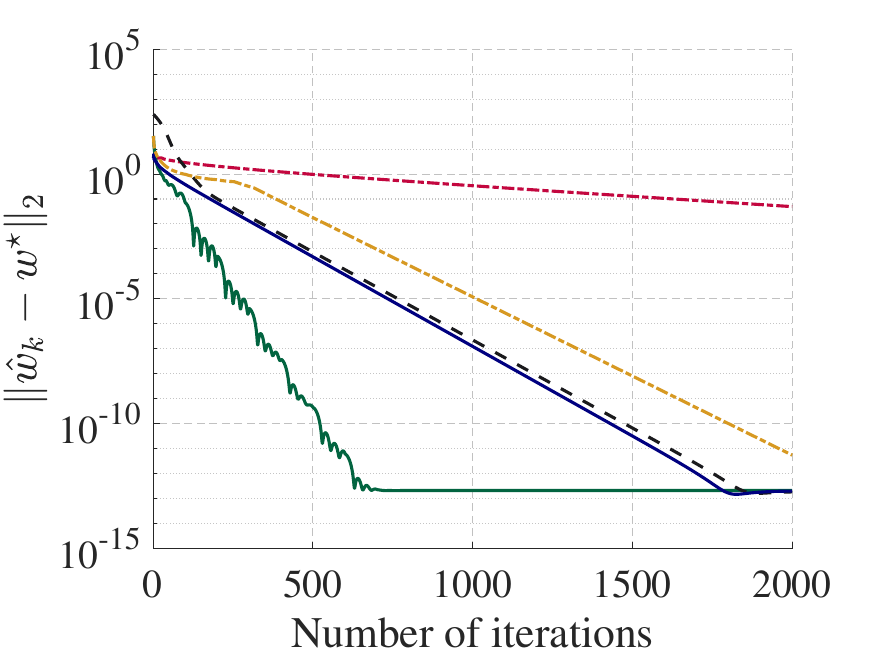}
    \end{minipage}
    \begin{minipage}[c]{.49\textwidth}
    \includegraphics[width=\textwidth]{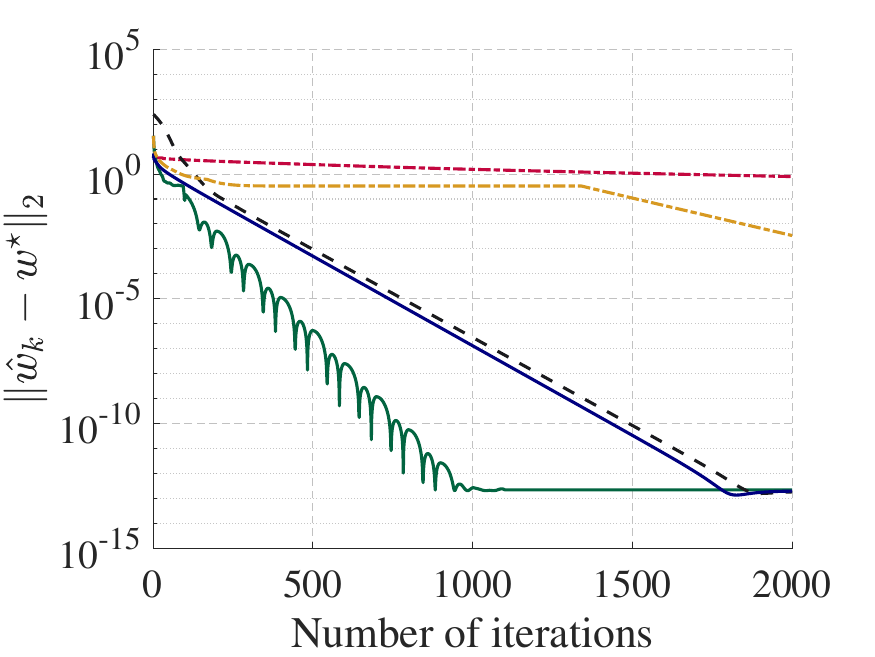}
    \end{minipage}
    \begin{minipage}[c]{.49\textwidth}
    \includegraphics[width=\textwidth]{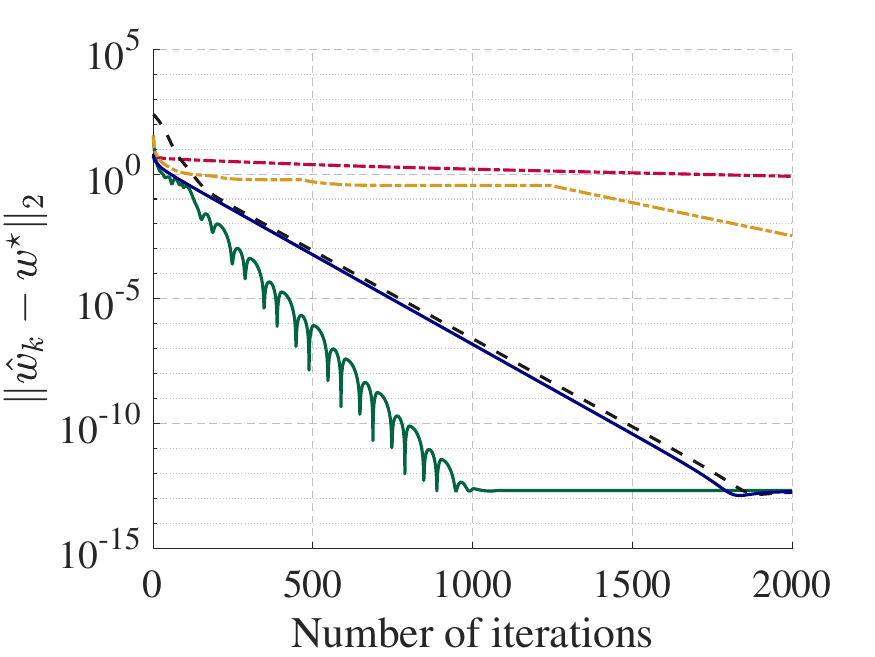}
    \end{minipage}
    \begin{minipage}[c]{.49\textwidth}
    \includegraphics[width=\textwidth]{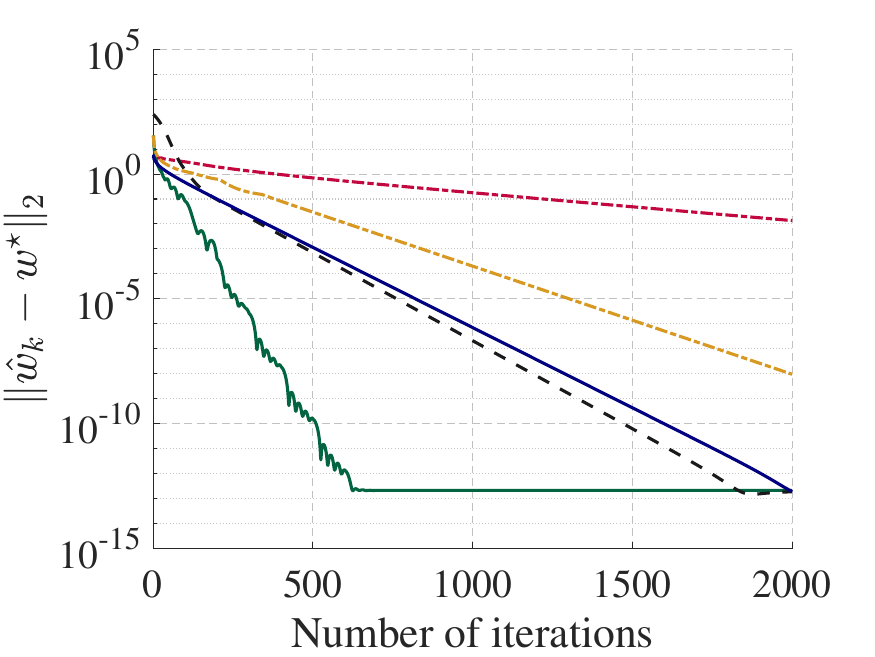}
    \end{minipage}
    \newline
    \centering{\small (b)}
    \end{minipage}
    \begin{minipage}[c]{.32\textwidth}
    \includegraphics[width=0.92\textwidth]{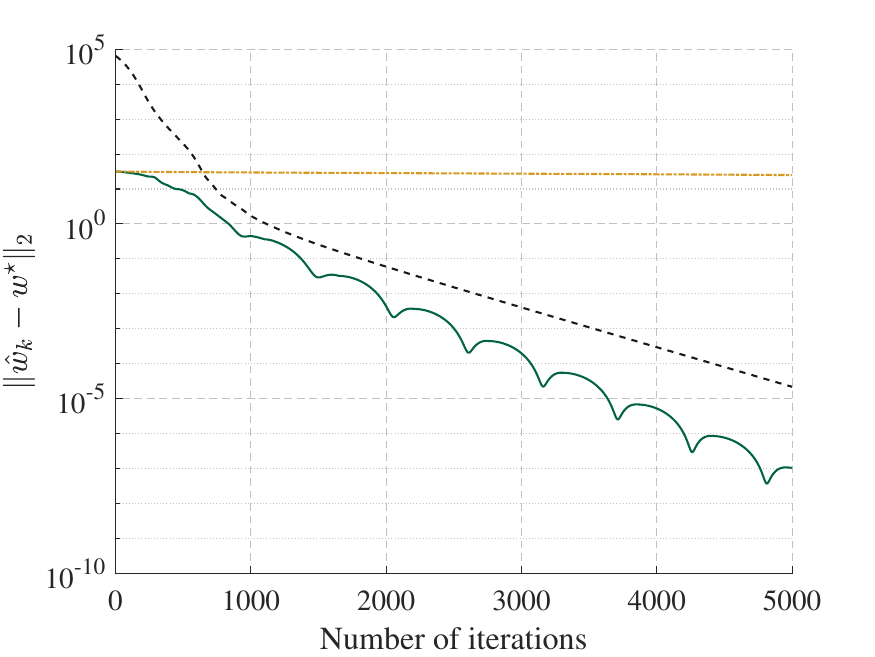}
    \newline
    \centering{\small (c)}
    \end{minipage}
    \caption{Convergence performance in terms of primal variable error $\|\hbw_k-\bbw^\star\|_2$ when recovering different synthetic and real graphs. (a) ER graphs with $N=200$ (top-left) and $N=400$ nodes (top-right); SBM graphs with $N=200$ (bottom-left), and $N=400$ nodes (bottom-right). (b) Four representative structural brain graphs with $N=66$ ROIs; Subject 1 (top-left),  Subject 2 (top-right), Subject 4 (bottom-left), and Subject 6 (bottom-right). (c) Minnesota road network with $N=2642$ intersections. In all cases, the proposed FDPG method converges faster to $\bbw^\star$ than state-of-the-art graph learning algorithms.}
    \label{fig_results}
\end{figure*}

Here we test the proposed fast dual PG (FDPG) algorithm for learning random and real-world graphs from simulated signals. The merits of the formulation \eqref{eq:kalofolias} in terms of recovering high-quality graphs have been well documented; see e.g.,~\cite{kalofolias16,kalofolias2019iclr,mateos19,dong2019learning} and references therein. For this reason, the numerical experiments that follow will exclusively focus on algorithmic performance, with no examination of the quality of the optimal solution $\bbw^\star$ that defines the learnt graph. In all ensuing test cases, we search for the best regularization parameters $\alpha,\beta$ in terms of graph recovery performance, adopting the edge-detection F-measure as criterion.
We compare Algorithm \ref{A:accelerated} to other state-of-the-art methods such as PD~\cite{kalofolias16}, PG~\cite{saboksayr21eusipco_ogl}, and linearized ADMM~\cite{wang2021}. We also consider the non-accelerated dual PG (DPG) method that is obtained from Algorithm \ref{A:accelerated} when $t_{k}\equiv 1$ for all $k\geq 1$.  For FDPG we implemented customary fixed-interval restarts of the momentum term in Algorithm \ref{A:accelerated}; see also~\cite{odonoghue2015restarts} for adaptive restart rules. Moreover, the ADMM parameters and PD step-size are tuned to yield the best possible convergence rate. Implementation details can be found in the publicly available code\footnote{\scriptsize{\texttt{\url{http://www.ece.rochester.edu/~gmateosb/code/FDPG.zip}}}.}, which can be used to generate all plots in Fig. \ref{fig_results}.\vspace{-0.2cm}


\subsection{Random graphs}\label{Ss:random_graphs}


We generate ground-truth graphs as draws from the Erd\H{o}s-R\'enyi (ER) model (edge formation probability $p=0.1$) with $N = 200$ and $400$ nodes, as well as from the $2$-block Stochastic Block Model (SBM) with the same number of nodes, and connection probability $p_1=0.3$ for nodes in the same community and $p_2=0.05$ for nodes in different blocks. We simulate $P=1000$ i.i.d. graph signals $\bbx_p \sim \ccalN\left( \bbzero, \bbL^{\dag}+\sigma_{e}^2 \bbI_N \right)$, where $\sigma_{e} = 0.1$ represents the noise level and $\bbL$ is the Laplacian of the ground-truth random graph.  For a graph-based factor analysis model justifying this approach to smooth signal generation, see e.g.,~\cite{dong16}. We compare the convergence performance of the aforementioned methods through the evolution of the primal variable error $\| \hbw_k - \bbw^{\star}\|_{2}$. To obtain $\bbw^{\star}$ for the chosen $\alpha$ and $\beta$, we ran the PD method for $50000$ iterations. The results of these comparisons are illustrated in Fig.~\ref{fig_results} (a). Apparently, the proposed FDPG algorithm markedly outperforms all other methods in terms of convergence rate, uniformly across graph model classes and number of nodes. Here, convergence to the largest graphs takes less iterations than for $N=200$.


\subsection{Brain and road networks}\label{Ss:real_graphs}


We first focus on recovering the topology of $6$ unweighted structural brain graphs~\cite{hagmann2008mapping}, all with $N=66$ regions of interest (ROIs) and whose edges connect ROIs with non-trivial density of axonal bundles; see also~\cite{segarra16ssp} for additional details. 
For a larger-scale experiment, we adopt the Minnesota road network which is an unweighted and undirected graph with $N = 2642$ intersections~\cite{davis2011}. In both cases, we generated synthetic smooth signals over the real topologies using the generative model in Section \ref{Ss:random_graphs}. The high value of $N$ renders the ADMM's 3-D parameter search a significantly time consuming operation.  Hence, for the Minnesota road network experiment, we only focus on the proposed (F)DPG methods and the PD algorithm in~\cite{kalofolias16}. 

Fig.~\ref{fig_results} (b) depicts the convergence results for the structural brain networks of $4$ representative subjects. Once more, in all cases the FDPG method is faster, but for these smaller graphs the performance gap appears to narrow. The gains can also be quantified in terms of wall-clock time. For instance, for Subject $6$ the time in seconds for the algorithms to reach a suboptimality of $10^{-8}$ are: $0.021$s for FDPG, $0.092$s for PD, $0.071$s for ADMM and $0.081$s for DPG.  Results for the Minnesota road network are depicted in Fig.~\ref{fig_results} (c), where the superiority of the proposed method is also apparent.


\section{Conclusion}\label{S:conclusions}


We developed a fast and scalable algorithm to learn the graph structure of signals subject to a smoothness prior. Leveraging this cardinal property of network data is central to various statistical learning tasks, such as graph smoothing and semi-supervised node classification.  We brought to bear a fast dual-based PG method to derive lightweight graph-learning iterations that come with global convergence rate guarantees. The merits of the proposed algorithm are showcased via experiments using several random and real-world graphs.


\newpage

\bibliographystyle{IEEEtran}
%
\bibliography{refs}

\begin{thebibliography}{10}
\providecommand{\url}[1]{#1}
\csname url@samestyle\endcsname
\providecommand{\newblock}{\relax}
\providecommand{\bibinfo}[2]{#2}
\providecommand{\BIBentrySTDinterwordspacing}{\spaceskip=0pt\relax}
\providecommand{\BIBentryALTinterwordstretchfactor}{4}
\providecommand{\BIBentryALTinterwordspacing}{\spaceskip=\fontdimen2\font plus
\BIBentryALTinterwordstretchfactor\fontdimen3\font minus
  \fontdimen4\font\relax}
\providecommand{\BIBforeignlanguage}[2]{{%
\expandafter\ifx\csname l@#1\endcsname\relax
\typeout{** WARNING: IEEEtran.bst: No hyphenation pattern has been}%
\typeout{** loaded for the language `#1'. Using the pattern for}%
\typeout{** the default language instead.}%
\else
\language=\csname l@#1\endcsname
\fi
#2}}
\providecommand{\BIBdecl}{\relax}
\BIBdecl

\bibitem{kolaczyk09}
E.~D. Kolaczyk, \emph{Statistical Analysis of Network Data: Methods and
  Models}.\hskip 1em plus 0.5em minus 0.4em\relax New York, NY:
  Springer\hyp{}Verlag, 2009.

\bibitem{ortega18}
A.~{Ortega}, P.~{Frossard}, J.~Kova\u{c}evi\'{c}, J.~M.~F. {Moura}, and
  P.~{Vandergheynst}, ``Graph signal processing: Overview, challenges, and
  applications,'' \emph{Proc. IEEE}, vol. 106, no.~5, pp. 808--828, 2018.

\bibitem{shuman13}
D.~I. {Shuman}, S.~K. {Narang}, P.~{Frossard}, A.~{Ortega}, and
  P.~{Vandergheynst}, ``The emerging field of signal processing on graphs:
  Extending high-dimensional data analysis to networks and other irregular
  domains,'' \emph{IEEE Signal Process. Mag.}, vol.~30, no.~3, pp. 83--98,
  2013.

\bibitem{SandryMouraSPG_TSP13}
A.~Sandryhaila and J.~M.~F. Moura, ``Discrete signal processing on graphs,''
  \emph{IEEE Trans. Signal Process.}, vol.~61, no.~7, pp. 1644--1656, Apr.
  2013.

\bibitem{dempster_cov_selec}
A.~P. Dempster, ``Covariance selection,'' \emph{Biometrics}, vol.~28, no.~1,
  pp. 157--175, 1972.

\bibitem{yuanlin2007}
M.~Yuan and Y.~Lin, ``Model selection and estimation in the {G}aussian
  graphical model,'' \emph{Biometrika}, vol.~94, no.~1, pp. 19--35, 2007.

\bibitem{glasso2008}
J.~Friedman, T.~Hastie, and R.~Tibshirani, ``Sparse inverse covariance
  estimation with the graphical lasso,'' \emph{Biostatistics}, vol.~9, no.~3,
  pp. 432--441, 2008.

\bibitem{Lake10discoveringstructure}
B.~M. Lake and J.~B. Tenenbaum, ``Discovering structure by learning sparse
  graphs,'' in \emph{Annual Cognitive Sc. Conf.}, 2010, pp. 778 -- 783.

\bibitem{egilmez2017jstsp}
H.~E. Egilmez, E.~Pavez, and A.~Ortega, ``Graph learning from data under
  {L}aplacian and structural constraints,'' \emph{IEEE J. Sel. Topics Signal
  Process.}, vol.~11, no.~6, pp. 825--841, Sep. 2017.

\bibitem{pavez2018tsp}
E.~Pavez, H.~E. Egilmez, and A.~Ortega, ``Learning graphs with monotone
  topology properties and multiple connected components,'' \emph{IEEE Trans.
  Signal Process.}, vol.~66, no.~9, pp. 2399--2413, May 2018.

\bibitem{kumar2020jmlr}
S.~Kumar, J.~Ying, J.~V. de~M.~Cardoso, and D.~P. Palomar, ``A unified
  framework for structured graph learning via spectral constraints,'' \emph{J.
  Mach. Learn. Res.}, vol.~21, no.~22, pp. 1--60, 2020.

\bibitem{segarra2016topoidTSP16}
S.~Segarra, A.~Marques, G.~Mateos, and A.~Ribeiro, ``Network topology inference
  from spectral templates,'' \emph{IEEE Trans. Signal Inf. Process. Netw.},
  vol.~3, no.~3, pp. 467--483, Aug. 2017.

\bibitem{pasdeloup2016inferenceTSIPN16}
B.~Pasdeloup, V.~Gripon, G.~Mercier, D.~Pastor, and M.~G. Rabbat,
  ``Characterization and inference of graph diffusion processes from
  observations of stationary signals,'' \emph{IEEE Trans. Signal Inf. Process.
  Netw.}, vol.~4, no.~3, pp. 481--496, 2018.

\bibitem{rasoul20}
R.~Shafipour and G.~Mateos, ``Online topology inference from streaming
  stationary graph signals with partial connectivity information,''
  \emph{Algorithms}, vol.~13, no.~9, pp. 1--19, Sep. 2020.

\bibitem{kalofolias16}
V.~Kalofolias, ``How to learn a graph from smooth signals,'' in \emph{Artif.
  Intel. and Stat. (AISTATS)}, 2016, pp. 920--929.

\bibitem{dong16}
X.~{Dong}, D.~{Thanou}, P.~{Frossard}, and P.~{Vandergheynst}, ``Learning
  {L}aplacian matrix in smooth graph signal representations,'' \emph{IEEE
  Trans. Signal Process.}, vol.~64, no.~23, pp. 6160--6173, 2016.

\bibitem{kalofolias17}
V.~{Kalofolias}, A.~{Loukas}, D.~{Thanou}, and P.~{Frossard}, ``Learning time
  varying graphs,'' in \emph{IEEE Intl. Conf. Acoust., Speech and Signal
  Process. (ICASSP)}, 2017, pp. 2826--2830.

\bibitem{kalofolias2019iclr}
V.~Kalofolias and N.~Perraudin, ``Large scale graph learning from smooth
  signals,'' in \emph{Int. Conf. Learning Representations (ICLR)}, 2019.

\bibitem{sundeep_icassp17}
S.~P. Chepuri, S.~Liu, G.~Leus, and A.~O. Hero, ``Learning sparse graphs under
  smoothness prior,'' in \emph{IEEE Intl. Conf. Acoust., Speech and Signal
  Process. (ICASSP)}, Mar. 2017, pp. 6508--6512.

\bibitem{mike_icassp17}
M.~G. Rabbat, ``Inferring sparse graphs from smooth signals with theoretical
  guarantees,'' in \emph{IEEE Intl. Conf. Acoust., Speech and Signal Process.
  (ICASSP)}, Mar. 2017, pp. 6533--6537.

\bibitem{sardellitti19}
S.~{Sardellitti}, S.~{Barbarossa}, and P.~{Di Lorenzo}, ``Graph topology
  inference based on sparsifying transform learning,'' \emph{IEEE Trans. Signal
  Process.}, vol.~67, no.~7, pp. 1712--1727, 2019.

\bibitem{berger2020graphlearning}
P.~{Berger}, G.~{Hannak}, and G.~{Matz}, ``Efficient graph learning from noisy
  and incomplete data,'' \emph{IEEE Trans. Signal Inf. Process. Netw.}, vol.~6,
  pp. 105--119, 2020.

\bibitem{bars19}
B.~{Le Bars}, P.~{Humbert}, L.~{Oudre}, and A.~{Kalogeratos}, ``Learning
  {L}aplacian matrix from bandlimited graph signals,'' in \emph{IEEE Intl.
  Conf. Acoust., Speech and Signal Process. (ICASSP)}, 2019, pp. 2937--2941.

\bibitem{mateos19}
G.~{Mateos}, S.~{Segarra}, A.~G. {Marques}, and A.~{Ribeiro}, ``Connecting the
  dots: Identifying network structure via graph signal processing,'' \emph{IEEE
  Signal Process. Mag.}, vol.~36, no.~3, pp. 16--43, 2019.

\bibitem{dong2019learning}
X.~{Dong}, D.~{Thanou}, M.~{Rabbat}, and P.~{Frossard}, ``Learning graphs from
  data: A signal representation perspective,'' \emph{IEEE Signal Process.
  Mag.}, vol.~36, no.~3, pp. 44--63, 2019.

\bibitem{giannakis18}
G.~B. Giannakis, Y.~Shen, and G.~V. Karanikolas, ``Topology identification and
  learning over graphs: Accounting for nonlinearities and dynamics,''
  \emph{Proc. IEEE}, vol. 106, no.~5, pp. 787--807, 2018.

\bibitem{beck2014}
A.~Beck and M.~Teboulle, ``A fast dual proximal gradient algorithm for convex
  minimization and applications,'' \emph{Operations Research Letters}, vol.~42,
  no.~1, pp. 1--6, 2014.

\bibitem{saboksayr21eusipco_ogl}
S.~S. Saboksayr, G.~Mateos, and M.~Cetin, ``Online graph learning under
  smoothness priors,'' in \emph{European Signal Process. Conf. (EUSIPCO)},
  Dublin, Ireland, 2021 (accepted; see also arXiv:2103.03762 [cs.LG]).

\bibitem{wang2021}
X.~Wang, C.~Yao, H.~Lei, and A.~M.-C. So, ``An efficient alternating direction
  method for graph learning from smooth signals,'' in \emph{IEEE Intl. Conf.
  Acoust., Speech and Signal Process. (ICASSP)}, Toronto, Canada, 2021, pp.
  5380--5384.

\bibitem{marques20}
A.~G. Marques, S.~Segarra, and G.~Mateos, ``Signal processing on directed
  graphs: The role of edge directionality when processing and learning from
  network data,'' \emph{IEEE Signal Process. Mag.}, vol.~37, no.~6, pp.
  99--116, 2020.

\bibitem{zhou04}
D.~Zhou and B.~Sch{\"o}lkopf, ``A regularization framework for learning from
  graph data,'' in \emph{Int. Conf. Mach. Learning (ICML)}, 2004.

\bibitem{chepuri17}
S.~P. {Chepuri}, S.~{Liu}, G.~{Leus}, and A.~O. {Hero}, ``Learning sparse
  graphs under smoothness prior,'' in \emph{IEEE Intl. Conf. Acoust., Speech
  and Signal Process. (ICASSP)}, 2017, pp. 6508--6512.

\bibitem{weiyu2018tsp}
W.~Huang, A.~G. Marques, and A.~R. Ribeiro, ``Rating prediction via graph
  signal processing,'' \emph{IEEE Trans. Signal Process.}, vol.~66, no.~19, pp.
  5066--5081, 2018.

\bibitem{beck18}
A.~Beck, \emph{First\hyp{}order Methods in Optimization}.\hskip 1em plus 0.5em
  minus 0.4em\relax Philadelphia, PA: Society for Industrial and Applied
  Mathematics, 2018.

\bibitem{beck2009}
A.~Beck and M.~Teboulle, ``A fast iterative shrinkage-thresholding algorithm
  for linear inverse problems,'' \emph{SIAM J. Img. Sci.}, vol.~2, no.~1, p.
  183–202, Mar. 2009.

\bibitem{boyd14}
N.~Parikh and S.~Boyd, ``Proximal algorithms,'' \emph{Foundations and Trends in
  optimization}, vol.~1, no.~3, p. 127–239, 2014.

\bibitem{odonoghue2015restarts}
B.~O'Donoghue and E.~Candes, ``Adaptive restart for accelerated gradient
  schemes,'' \emph{Foundations of Computational Mathematics}, vol.~15, pp.
  715--732, 04 2015.

\bibitem{hagmann2008mapping}
P.~Hagmann, L.~Cammoun, X.~Gigandet, R.~Meuli, C.~J. Honey, V.~J. Wedeen, and
  O.~Sporns, ``Mapping the structural core of human cerebral cortex,''
  \emph{PLoS Biol}, vol.~6, no.~7, p. e159, 2008.

\bibitem{segarra16ssp}
S.~Segarra, A.~G. Marques, G.~Mateos, and A.~Ribeiro, ``Network topology
  identification from spectral templates,'' in \emph{IEEE Wrkshp. Statistical
  Signal Process. (SSP)}, 2016, pp. 1--5.

\bibitem{davis2011}
T.~A. Davis and Y.~Hu, ``The {U}niversity of {F}lorida {S}parse {M}atrix
  {C}ollection,'' \emph{ACM Trans. Math. Softw.}, vol.~38, no.~1, Dec. 2011.

\end{thebibliography}


\newpage


\section*{Supplementary Material}\label{S:supp_material}



\subsection*{Proof of Lemma \ref{lemma:Lipschitz_gradient}}\label{App:proof_Lipschitz_gradient}


A couple preliminary calculations are required to derive an explicit expression for the Lipschitz constant of $F(\bblambda)$.\vspace{-5pt}
\begin{mylemma}\label{lemma:strong_convexity}
The function $f(\bbw):=\mbI\left\{\bbw\succeq\mathbf{0}\right\} + 2\bbw^{\top}\bbe+\beta\|\bbw\|_2^2$ is strongly convex with constant $\sigma:=2\beta>0$.
\end{mylemma}
\begin{myproof}
The strong convexity of $f$ with parameter $\sigma = 2\beta>0$ follows because
\begin{equation*}
	f(\bbw) - \frac{\sigma}{2}\|\bbw\|^2 = \mbI\left\{\bbw\succeq\mathbf{0}\right\} + 2\bbw^{\top}\bbe
\end{equation*} 
is a convex function.
\end{myproof} \vspace{-5pt}
\begin{mylemma}\label{lemmma:spectral_norm}
Going back to \eqref{eq:kalofolias_vec}, recall $\bbS\in\{0,1\}^{N\times N(N-1)/2}$ defined so that $\bbd=\bbW\bbone=\bbS\bbw$. Then, $\|\bbS\|_2=\sqrt{2(N-1)}$.
\end{mylemma}

\begin{myproof}
Because $\bbS$ maps the upper-triangular adjacency matrix entries in $\bbw$ to the degree sequence $\bbd$, then $\bbS$ has $N-1$ ones in each row while all other entries are zero. Hence, the diagonal entries of $\bbS\bbS^{\top}$ are all $N-1$ and the off-diagonal entries are equal to $1$. The eigenvalues $\lambda$ of $\bbS\bbS^{\top}=(N-2)\bbI  + \bbone\bbone^{\top}$ are the roots of the characteristic polynomial
\begin{align*}
		\text{det}\left( \bbS\bbS^{\top} - \lam\bbI \right) &= \text{det}\left( (N-2)\bbI  + \bbone\bbone^{\top} - \lam \bbI \right)\\
		&= \text{det}( \underbrace{(N-2-\lam)\bbI}_{:=\bbQ}  + \bbone\bbone^{\top})\\
		&= \text{det}(\bbQ) + \bbone^{\top}\text{adj}(\bbQ)\bbone\\
		&= \prod_{i=1}^{N} Q_{ii} + \sum_{j=1}^{N}\prod_{i\neq j} Q_{ii}\\
		&= (N-2-\lam)^{N} + N(N-2-\lam)^{N-1}\\
		&= (2N-2-\lam)(N-2-\lam)^{N-1}=0.
\end{align*}
To obtain the third equality we leveraged the Sherman-Morrison formula, where $\textrm{adj}(\bbQ)$ stands for the adjugate matrix of $\bbQ$.
From the final factorization of the polynomial, the eigenvalues are $ 2(N-1)=\lam_1>\lam_2=\dots=\lam_N = N-2$. Because $\|\bbS\|_2=\sqrt{\lam_1}$, the result follows.
\end{myproof}

Since $f$ is strongly convex (with constant $\sigma$), by virtue of~\cite[Lemma 3.1]{beck2014} the function $F(\bblambda):=\max_{\bbw}\left\{\langle \bbS^\top \bblambda,\bbw\rangle-f(\bbw)\right\}$ is continuously differentiable and it has a Lipschitz continuous gradient with constant $L:=\frac{\|\bbS\|_2^2}{\sigma}$. From the expressions for $\sigma$ and $\|\bbS\|_2$ in Lemmata \ref{lemma:strong_convexity} and \ref{lemmma:spectral_norm}, the result follows.\hfill$\square$


\subsection*{Proof of Proposition \ref{prop:iterations}}\label{App:proof_prop_iterations}


Leveraging the result in~\cite[Lemma 3.2]{beck2014}, it follows that for all $k\geq 1$, the dual variable update iteration in \eqref{eq:FISTA_prox} can be equivalently rewritten as $\bblambda_k=\bbomega_k-L^{-1}(\bbS\bar{\bbw}_k-\bbu_k)$, with
\begin{align} 	
	\bar{\bbw}_k={}& \argmax_\bbw\left\{\langle \bbS^\top \bbomega_k,\bbw\rangle-f(\bbw)\right\},\label{eq:barw_update_raw}\\
	\bbu_k={}&\textbf{prox}_{Lg}\left(\bbS\bar{\bbw}_k-L\bbomega_k\right).\label{eq:u_update_raw}
\end{align}	
Starting with \eqref{eq:barw_update_raw}, we have from the definition of $f$ that
\begin{align*} 	
	\bar{\bbw}_k={}& \argmin_\bbw\left\{\mbI\left\{\bbw\succeq\mathbf{0}\right\} + \beta\|\bbw\|_2^2-\langle \bbS^\top \bbomega_k-2\bbe,\bbw\rangle\right\}\\
	={}&\argmin_\bbw\left\{\mbI\left\{\bbw\succeq\mathbf{0}\right\} +\frac{1}{2}\left\|\bbw-\frac{\bbS^\top \bbomega_k-2\bbe}{2\beta}\right\|_2^2 \right\}\\
	={}& \textbf{prox}_{\mbI\left\{\bbw\succeq\mathbf{0}\right\}}\left(\frac{\bbS^\top \bbomega_k-2\bbe}{2\beta}\right)\\
	={}&\max\left(\mathbf{0},\frac{\bbS^\top\bbomega_k-2\bbe}{2\beta}\right)
\end{align*}	
as desired [cf. \eqref{eq:barw_update}]. The last equality follows from the fact that the proximal operator of $\mbI\left\{\bbw\succeq\mathbf{0}\right\}$ is the projection onto the non-negative orthant $\bbw\succeq\mathbf{0}$.

To arrive at the update of $\bbu_k$ in \eqref{eq:u_update}, it suffices to start from \eqref{eq:u_update_raw} and recall that the proximal operator of $Lg(\bbw)=-L\alpha \bbone^{\top} \log \left( \bbw \right)$ is given by (see e.g.,~\cite{kalofolias16} and~\cite[Proposition 2]{wang2021})
\begin{equation*}
\textbf{prox}_{Lg}(\bbw)=\frac{\bbw  + \sqrt{\bbw^2 + 4\alpha L\bbone}}{2},
\end{equation*}
where the square and square root are understood to be taken element-wise. Evaluating the proximal operator at $\bbS\bar{\bbw}_k-L\bbomega_k$, the result follows.\hfill$\square$


\subsection*{Dual suboptimality}\label{App:dual_suboptimality}


\begin{figure}[t!]
	\centering
	\includegraphics[width=\linewidth]{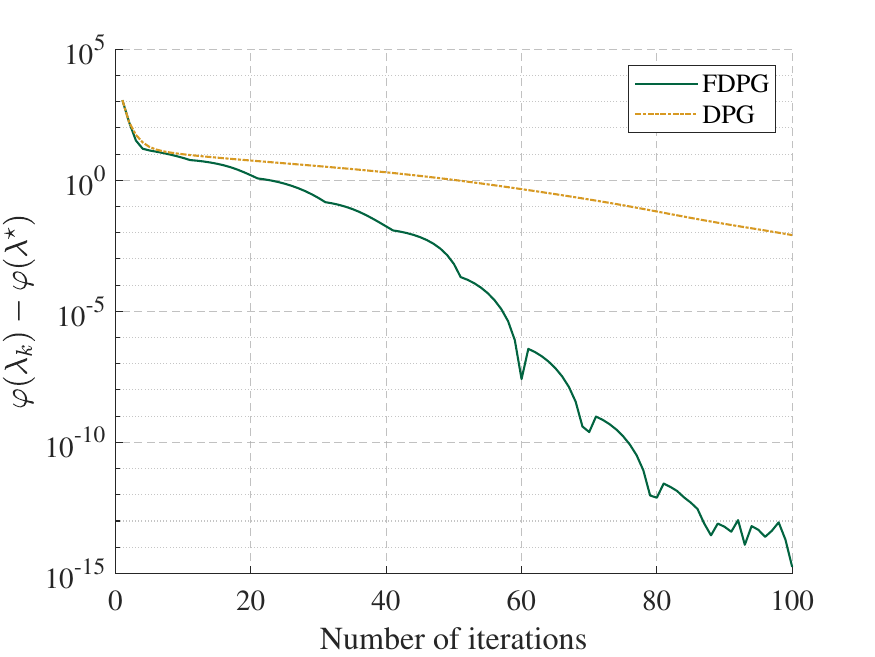}
	\caption{Convergence performance in terms of dual suboptimality $\varphi(\bblambda_k)-\varphi(\bblambda^\star)$, when recovering the SBM graph with $N=200$ nodes described in Section \ref{Ss:random_graphs}. As expected, the FDPG graph learning algorithm converges markedly faster than its non-accelerated counterpart.}
	\label{fig:sim_dual_obj}
\end{figure}

Recall one of the test cases in Section \ref{Ss:random_graphs}, where the goal was to recover a $2$-block Stochastic Block Model (SBM) with $N=200$ nodes from $P=1000$ synthetically-generated smooth signals. In Fig.~\ref{fig:sim_dual_obj} we depict the evolution of the dual suboptimality $\varphi(\bblambda_k)-\varphi(\bblambda^\star)$ for the FDPG (Algorithm \ref{A:accelerated}) and DPG iterations.  As expected, the proposed FDPG solver markedly outperforms its non-accelerated counterpart in terms of convergence rate. Similar behavior can be observed for random graphs drawn from the Erd\H{o}s-R\'enyi (ER) model and for different values of $N$ and $P$. The corresponding plots are not included due to lack of space.

\end{document}